\documentclass[preprint,12pt]{article}

\usepackage{authblk}
\usepackage{graphicx}
\usepackage{amssymb}

\usepackage{amsmath}
\usepackage{url}

\usepackage{amsthm}
\theoremstyle{definition} 
\newtheorem{definition}{Definition}[section]
\theoremstyle{remark} 
\newtheorem{remark}{Remark}[section]
\theoremstyle{example} 
\newtheorem{example}{Example}[section]

\usepackage{color}

\begin{document}

\title{A consensus set for the aggregation of partial rankings: the case of the Optimal Set of Bucket Orders Problem}

\date{}

\author[1]{Juan A. Aledo} 

\affil[1]{ \small
        JuanAngel.Aledo@uclm.es, ORCID 0000-0003-1786-8087 \\
        
        Dep. Matemáticas, Universidad de Castilla-La Mancha, Albacete, 02071, Spain.
}

\author[2]{Jos\'e A. G\'amez}

\affil[2]{
    Jose.Gamez@uclm.es, ORCID 0000-0003-1188-1117 \\
    
   Dep. Sistemas Informáticos, Universidad de Castilla-La Mancha, Albacete, 02071. Spain.
}

\author[3,4]{Alejandro Rosete } 

\affil[3]{
    Rosete@ceis.cujae.edu.cu, ORCID 0000-0002-4579-3556 \\
    
    Universidad Tecnológica de La Habana Jose Antonio Echeverría, Marianao, 19390, La Habana, Cuba.
}

\affil[4]{
    Avangenio S.R.L., 5ta B. esq. 6, La Habana, Cuba. 
}
            
\maketitle 

\newpage

\begin{abstract}
In rank aggregation problems (RAP), the solution is usually  a  consensus ranking that generalizes a set of input orderings. There are different variants that differ not only in terms of the type of rankings that are used as input and output, but also in terms of the objective function employed to evaluate the quality of the desired output ranking. In contrast, in some machine learning tasks (e.g. subgroup discovery) or multimodal optimization tasks, attention is devoted to obtaining several models/results  to account for the diversity in the input data or across the search landscape. Thus, in this paper we propose to provide, as the solution to an RAP, a set of rankings to better explain the preferences expressed in the input orderings. We exemplify our proposal through the Optimal Bucket Order Problem (OBOP), an RAP which consists in
finding a single consensus ranking (with ties) that generalizes a set of input rankings codified as a precedence matrix. To address this, we introduce the Optimal Set of Bucket Orders Problem (OSBOP), a generalization of the OBOP that aims to produce not a single ranking as output but a set of consensus rankings. 
Experimental results are presented to illustrate this proposal, showing how, by providing a set of consensus rankings, the fitness of the solution significantly improves with respect to the one of the original OBOP, without losing  comprehensibility. 
\end{abstract}


\section{Introduction}\label{sec:intro}

 Rankings are ordered lists of items generated by a human or machine-based voter. They find applications in a wide range of disciplines, from search engines that organize results by relevance, to sports leagues that establish team standings, and businesses that prioritize products or services. 
In scenarios where multiple rankings are provided (each reflecting distinct criteria or viewpoints), combining these into a single, representative ranking becomes a complex challenge. This task, known as ranking aggregation, involves reconciling the differences among individual lists, addressing conflicts such as varying priority orders, and balancing criteria to accurately represent the collective preferences or priorities. Effective ranking aggregation techniques are essential for producing fair and meaningful results in applications ranging from social choice theory and recommendation systems to meta-search engines and decision-making processes.

Real-world applications of rank aggregation arise in fields such as social choice theory, voting, consensus decision-making, hiring and recruitment processes, meta-search, seriation, web browsing patterns, and more. Many of these applications entail the ranking of sensitive information, such as political candidates or job applicants, which raises issues of bias and fairness that ranking algorithms must address. Recently, rank aggregation approaches have been developed to consider the existence of protected attributes like gender, race, or religion, resulting in groups that should be taken into account when generating the consensus ranking \cite{FairRAP:2024,FairnesRanking:2021}. The main idea in these cases is to incorporate fairness-based metrics that constrain the output, ensuring that a fair representation of each group appears within the top-k positions of the consensus ranking (e.g. proportional fairness \cite{Baruah:1996} or equal opportunity \cite{Hardt:2016}). In practice, existing approaches aim to produce a {\em fair} consensus ranking, which meets fairness requirements while minimizing the distance  to the consensus ranking that would be obtained without fairness considerations \cite{Balestra2024,Kuhlman:2020}  or implement postprocessing techniques on the obtained {\em unfair} consensus \cite{Gorantla:2021}. Additionally, strategic weight assignments in the aggregation process have been employed to mitigate recommendation bias issues \cite{DONG2019}, along with methods that account for the attentive influence of preference users \cite{Yu2023}. All these approaches produce a single ranking as a response, which summarizes the precedences provided in the input according to diverse criteria.

Mathematically, a rank aggregation problem (RAP) consists in finding the consensus order $\pi_0$ of $n$ items which best aligns with a set of preferences $\Pi$ for those items (for instance a set of rankings, a set of pairwise preferences or a precedence matrix) according to a specific criterion or distance measure $d$, i.e.
\[
\pi_0={\rm argmin}_{\pi} d(\Pi, \pi),
\]
where $\pi$ belongs to  a set of orderings of these items. There are many variants of RAP, which differ in terms of the types of rankings used as input and output, as well as in the objective function employed to evaluate the quality of the output ranking. In fact, the rankings involved, both in the input and output, may be either complete (ranking all $n$ items) or incomplete (ranking only a subset of the items) and may allow for ties (items between which there is no preference) or not. Furthermore,  a wide range of distance measures can be used to assess the quality of the output (Kendall's Tau Distance, Spearman’s Rank Correlation, Hamming Distance, etc.) \cite{Kumar:2010}. 

Typically, rank aggregation problems aim to identify a single consensus order. In contrast, some machine learning and optimization tasks, e.g., subgroup discovery \cite{SubgroupDiscovery:2011}, multiple solutions identification \cite{Kumar:2020} and multimodal optimization \cite{Yu2010} tasks, focus on obtaining multiple models or results to better capture the diversity present in the input data or the search landscape. Accordingly, in this paper, we propose providing a set of rankings as the solution to  an RAP to more effectively reflect the range of preferences within the input orderings. Specifically, for an RAP with input set of preferences $\Pi$, a distance measure $d$, and a natural number $b\geq 2$, we can define the generalized RAP, which consists in finding $b$ rankings $\pi_1,\pi_2,\dots \pi_b$  and $b$ weights $w^1,w^2,\dots w^b\in [0,1]$, $\sum_{k=1..b} w^k =1$, achieving the best fitness value
\[
\sum_{k=1}^b w^k d(\Pi,\pi_k)
\]

To illustrate this proposal, we deal with the Optimal Bucket Order Problem (OBOP) \cite{Gionis2006,Ukkonen2009}, a distance-based rank aggregation problem where the input is a pair order matrix $C$ of order $n$ which (usually) codifies the precedence relations between $n$ items, and the output is a bucket order, i.e., a complete ranking with ties (see Section \ref{sec:OBOP}). As explained above, we extend this ranking aggregation framework to account for the underlying presence of different groups within the set of voters or in their stated preferences (pair order matrix). Thus, we define the Optimal Set of Bucket Orders Problem (OSBOP) which, instead of constraining the consensus ranking with predetermined fairness criteria,  aims to produce a set of bucket orders, each assigned a weight to reflect its significance. The idea is that each learned bucket order can represent the consensus ranking for a community within the population from which the input preferences were gathered, with the goal that the weighted linear combination of these bucket order matrices minimizes the distance to the input pair order matrix (see Section \ref{sec:OSBOP}). Note that our approach differs from performing a clustering process to identify groups (e.g. \cite{Kamishima:2006}) and then solving an OBOP for each group individually. The distinction is twofold: first, in our proposal, weights and bucket orders are sought simultaneously to minimize the distance to the input pair order matrix. Second, if the input to our problem is directly a precedence matrix instead of a set of rankings, then the clustering approach becomes infeasible. With this approach in mind, our main contributions are:

\begin{itemize}
\item The definition of the Optimal Set of Bucket Orders Problem (OSBOP), where a set of bucket orders and their associated weights are obtained instead of a single bucket order. The OSBOP can be seen as a generalization of the OBOP.

\item A discussion of this new problem is presented to provide insight, including its relationship to the utopian matrix defined in \cite{Aledo2017Utopia}, along with illustrative examples that compare the OSBOP with the OBOP.

\item The introduction of a simple metaheuristic technique for addressing the OSBOP.

\item A repository with the implementation of the algorithms, to ensure the reproducibility of the results.
\end{itemize}

The remainder of this paper is organized as follows. Section \ref{sec:set-up} introduces background information on bucket orders and the OBOP. In Section \ref{sec:OSBOP}, we present the OSBOP as a generalization of the OBOP. Section \ref{sec:method} describes a simple metaheuristic method for addressing the OSBOP. Finally, Sections \ref{sec:experiments} and \ref{sec:conclusions} cover the experiments conducted and our conclusions, respectively.

\bigskip

\section{Set up}\label{sec:set-up}
\subsection{Bucket orders}
A  \emph{bucket order} $\mathcal{B}$ (a.k.a. \emph{partial ranking} or \emph{complete ranking with ties}) of $n$ elements is an ordered partition of $[[n]]=\{1,2,\dots,n\}$ \cite{Fagin2004,Gionis2006,Ukkonen2009}, i.e. a linear ordering of \emph{buckets} $B_1,B_2,\dots,B_k\subseteq[[n]]$, $1\leq k\leq n$, with $\cup_{i=1}^k B_i=[[n]]$ and $B_i \cap B_j = \emptyset,$ $i\neq j$. We denote by $\widetilde{\mathbb{S}}_n$ the set of bucket orders of the elements in $[[n]]$.

For two buckets $B_i, B_j$ in $\mathcal{B}$, if $B_i$ precedes $B_j$ according to the bucket order $\mathcal{B}$, we write $B_i\prec_\mathcal{B} B_j$. This relation also apply to the items included in the buckets. Specifically, if $u \in B_i$, $v \in B_j$ and  $B_i\prec_\mathcal{B} B_j$, then we write $u\prec_\mathcal{B} v$. On the other hand, if two items $u,v$ are \emph{tied} (i.e. they belong to the same bucket), we write $u \sim_{\mathcal{B}} v$.
   
 We represent a bucket order as a list of the items in $[[n]]$, where the items in the same bucket are separated by commas ($,$) and the bar ($|$)  indicates the separation between buckets. For example,  $1,3|2,4$ represents a bucket order of the elements in $[[4]]$ with two buckets: the first consisting of items 1 and 3, followed by the second containing items 2 and 4. This implies that 1 and 3 are considered equally preferred (tied), as are 2 and 4, with 1 and 3 being preferred over 2 and 4.

We will identify  a bucket order $\mathcal{B}$ with a square matrix $n\times n$ called \emph{bucket matrix} $B$ \cite{Gionis2006,Ukkonen2009}, where $B(u,v)=1$ if $u\prec_\mathcal{B} v$, $B(u,v)=0$ if $v\prec_\mathcal{B} u$, and $B(u,v)=0.5$ if $u \sim_{\mathcal{B}} v$. 
The elements in the main diagonal of $B$ are equal to $0.5$, and $B(u,v)+B(v,u)=1$ for all $u,v\in[[n]]$, $u\neq v$. On the other hand, a \emph{pair order matrix} $C$ of order $n$ is a square matrix $n\times n$ satisfying that $C(u,v)\in [0,1]$, $C(u,v)+C(v,u)=1$ and $C(u,u)=0.5$ for all $u,v\in[[n]]$. Note that, in general, a pair order matrix is not a bucket order matrix because the transitivity property typically fails to hold.

\subsection{The Optimal Bucket Order Problem (OBOP)} \label{sec:OBOP}
In this section, we briefly revise the  Optimal Bucket Order Problem (OBOP) (see \cite{Gionis2006,Ukkonen2009}). 
The OBOP is a distance-based rank aggregation problem whose input is a pair order matrix $C$ of order $n$ which (usually) codifies the precedence relations between the items in $[[n]]$, whereas the output is a bucket order in $\widetilde{\mathbb{S}}_n$.

Specifically, given a pair order matrix $C$ of order $n$, the OBOP consists in finding a bucket order $\mathcal{B}$, and more precisely a bucket matrix $B$, which minimizes the distance
\begin{equation}
D(B,C)=\sum_{u,v\in [[n]]} |B(u,v)-C(u,v)|.
\end{equation}

As the OBOP is an NP-Complete problem \cite{Gionis2006}, various heuristic and greedy approaches have been developed to address it (e.g. \cite{Aledo2018-ES-EJOR, Aledo2021,Ukkonen2009}).

In \cite{Aledo2017Utopia} the concept of \emph{ utopian matrix} for the OBOP was introduced as follows. Given a pair order matrix $C$ of order $n$, 
the associated \emph{utopian matrix} $U_C$ is the $n\times n$ matrix 
\[
U_C(u,v)=\Upsilon(C(u,v)), \qquad u,v\in [[n]],
\]
where
\[
\Upsilon(x)=\left\{
\begin{array}{rl}
1 \qquad &  \mbox{if $x > 0.75$} \\
0.5 \qquad &  \mbox{if $0.25\leq x \leq 0.75$} \\
0 \qquad &  \mbox{if $x < 0.25$}
\end{array}
\right.
\]
The idea behind the utopian matrix $U_C$ associated with the pair order matrix $C$ is to assign, according to the function $\Upsilon(C(u,v))$, the value $U_C(u,v)\in \{0,0.5,1\}$ (possible entries in a bucket matrix) that is closer to $C(u,v)$.  
We also define the \emph{utopia value} $u_C$ associated with $C$ as $u_C=D(U_C,C)$.  
The utopian matrix $U_C$ is a superoptimal solution for the OBOP, i.e. $D(B,C)\geq D(U_C,C)= u_C$ for whichever bucket matrix $B$.
In particular, when the utopian matrix is a bucket matrix (which is not the case, in general) it becomes the optimal solution to the OBOP.

\section{The Optimal Set of Bucket Orders Problem (OSBOP)}
\label{sec:OSBOP}

In this section, we formalize the Optimal Set of Bucket Orders Problem (OSBOP) as a generalization of the  OBOP.

Given $b$ bucket matrices $B_1,B_2,\dots, B_b$ and real numbers $w^1,w^2,\dots w^b\in [0,1]$, $\sum_{k=1}^b w^k =1$, we will denote by $\overline{B}[(B_1,\dots,B_b),(w^1,\dots,w^b)]$ the linear combination of the matrices $\{B_k\}_1^b$ according to the weights $\{w^k\}_1^b$, namely
    \begin{equation}\label{Bbar}
    \overline{B}[(B_1,\dots,B_b),(w^1,\dots,w^b)]=\sum_{k=1}^b w^k B_k.
    \end{equation}

\begin{definition}[OSBOP]
Given a pair order matrix $C$ and a positive integer $b$, the $b$-th Optimal Set of Bucket Orders Problem ({\rm OSBOP}$^b$) consists in finding $w^1,w^2,\dots w^b\in [0,1]$, $\sum_{k=1..b} w^k =1$,  and $b$ (different) bucket matrices $B_1,B_2,\dots, B_b$ that minimize the distance
\begin{eqnarray}
f_C(\overline{B}) & = & D(\overline{B}[(B_1,\dots,B_b),(w^1,\dots,w^b)],C)  \nonumber\\
& = &\sum_{u,v\in [[n]]} |\overline{B}[(B_1,\dots,B_b),(w^1,\dots,w^b)](u,v)-C(u,v)|. \label{fC}
\end{eqnarray}
\end{definition}

Thus, in contrast to the OBOP where the solution is a single bucket order, for the OSBOP the solution consists of a set of bucket orders and a vector of weights that express the relative importance of each bucket order in the set. 
The allowed number of bucket orders $b>0$ in the output defines different versions of the problem. Hence, we denote OSBOP$^b$ the specific OSBOP  where the number of allowed bucket orders is $b$. 
In particular, OSBOP$^1$ is simply the  OBOP. It is also worth noting that the set of solutions of OSBOP$^b$ is included within that of OSBOP$^{b+1}$, since assigning a weight equal to $0$ effectively excludes a particular bucket order from being considered in the solution. 

We will denote by OSBOP$_e^b$ the OSBOP$^b$ where uniform weights are considered, i.e. $w^k = 1/b, k=1, \dots, b$. 
Assigning equal values to all weights is analogous to requiring, in a clustering problem, that all clusters be of equal size, or that the branches of a decision tree remain balanced. These scenarios share a commonality: the overall model is treated as a composition of sub-models with equal importance. If it is not the case, the sub-models may vary in significance (number of elements in clustering, rules support in decision trees, weights in OSBOP) which often results in a higher-quality overall model. Consequently, such constraints depend on the user’s objectives. Finally, observe that the set of solutions of the OSBOP$^b_e$ is included within that of the OSBOP$^{b+1}$ by simply assigning one of the weights a value of 0.

 To illustrate this scenario, let us assume a simple situation where several people give their preference about four foods (1:chicken, 2:fish, 3:fruits, and 4:nuts), where 60\% of the people are meat-eaters that prefer chicken and fish, while the other 40\% are vegetarian that prefer fruit and nuts. Assume futher that the meat-eater's preference is represented by the ranking $1,2|3,4$  while the vegetarian's preference  is expressed as $3,4|1,2$. 
Then, the pair order matrix which codifies the users preferences is
\[
C=\left(
  \begin{array}{cccc}
0.5     & 0.5  &  0.6 & 0.6 \\
0.5     & 0.5 &  0.6 & 0.6  \\
0.4     & 0.4  &  0.5 & 0.5     \\
0.4     & 0.4  &  0.5 & 0.5     \\
  \end{array}
\right).
\]
 For this instance, the optimal solution to the OBOP  is the bucket order $1,2,3,4$ where all foods are tied, i.e. the bucket order matrix $B$ with $B(u,v)=0.5$ for all $u,v\in [[4]]$. Actually (see Section \ref{utopian matrix OSBOP}), $B$ is the utopian matrix associated with $C$ and so  is the optimal solution to the OPOB  for the preference matrix $C$. In particular, $D(B,C)=0.8$. This solution assumes that there are no specific food preferences for any user, which does not accurately reflect the real situation.

 However, the solution to the OSBOP$^2$ consists of the rankings $B_1=1,2|3,4$ with 
 $w^1=0.6$, and $B_2=3,4|1,2$ with $w^2=0.4$.  Namely, the matrix $\overline{B}[(B_1,B_2),(w^1,w^2)]$ coincides with the preference matrix $C$, being $D(\overline{B}[(B_1,B_2),(w^1,w^2)], C)=0$. In other words, the OSBOP approach faithfully reflects the set of preferences, avoiding the assumption of overall indifference inherent in the OBOP solution which does not align with reality.

Assume now that the preferences are more imbalanced, for instance, with 90\% of individuals being meat-eaters. Then the optimal solution to the OBOP is $1,2|3,4$  (again, it corresponds to the utopian matrix for this instance, with a distance of 0.8), which ignores the existence of a minority of vegetarians. 
On the other hand, the optimal solution to the OSBOP$^2$ is the same pair of rankings $B_1$ and $B_2$ described previously, now with weights $w^1=0.9$  and $w^2=0.1$. Once again, this solution accurately reflects the users preferences allowing for the expression of the minority vegetarians’s options.

\subsection{Size of the solution space}

\begin{figure}[htbp]
\centering
\includegraphics[width=0.8\textwidth]{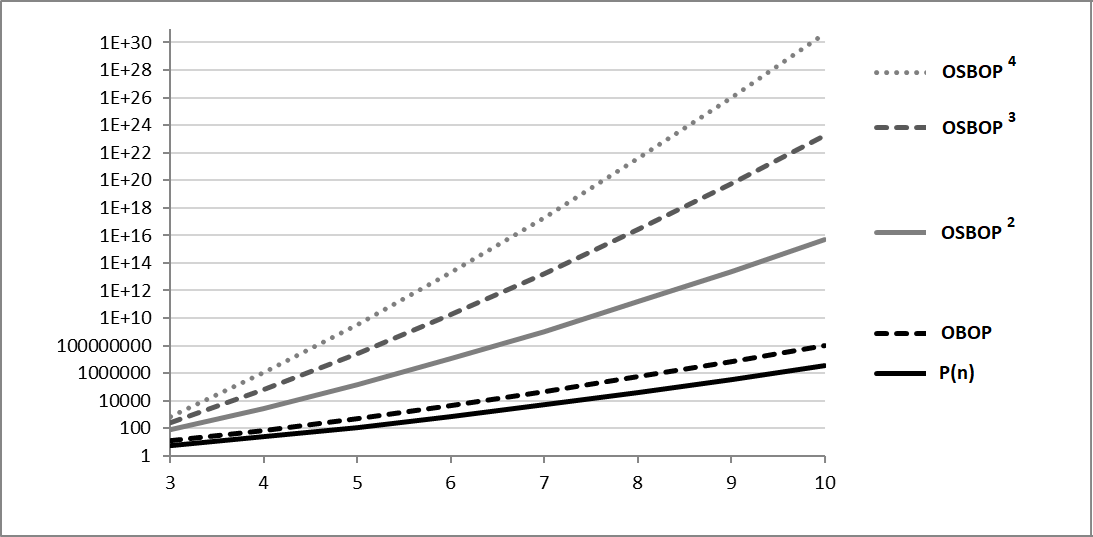}
\caption{\label{fig:SizeSpace}Size of the solution spaces for different variants of OSBOP.}
\end{figure}

The size of the solution space of the OSBOP$^b$ is 
\[
C^{|\widetilde{\mathbb{S}}_n|}_b=\binom{|\widetilde{\mathbb{S}}_n|}{b} =\frac{|\widetilde{\mathbb{S}}_n| !}{ b! (|\widetilde{\mathbb{S}}_n|-b)! },
\]
where the cardinal of $\widetilde{\mathbb{S}}_n$ is determined by the Fubini number $F(n)$ (see, for instance, \cite[Section 3]{Aledo2018-ES-EJOR} for the details). 
In the particular case of the OSBOP$^1$ (i.e. OBOP), the size of the solution space is $|\widetilde{\mathbb{S}}_n|$. In Figure \ref{fig:SizeSpace} we show, in polynomial scale, the sizes of the solution spaces of the OSBOP$^b$ for $1\leq b \leq 4$ and $2\leq n\leq 10$. 
We also show the number of permutations of $n$ elements, $P(n)=n!$, which corresponds to the size of the space of the OSBOP$^1$ when no ties are allowed in the output ranking.

\subsection{The utopian matrix for OSBOP}\label{utopian matrix OSBOP}

In order to extend the definition of utopian matrix to the OSBOP$_e^b$, $b\geq 2$, note that the entries of a linear combination of $b$ bucket matrices with weights $1/b$ are in the set
\[
\left\{0, \frac{1}{2b}, \dots, \frac{2b-1}{2b}, \frac{2b}{2b}=1 \right\}.
\]
Then, we define the \emph{$b$-th utopian matrix} $U_C^b$ of a pair order matrix $C$, $b\geq 2$, as
\[
U_C^b(u,v)=\Upsilon^b(C(u,v))
\]
where

\[
\Upsilon^b(x)=\left\{
\begin{array}{rl}
1 & \mbox{if $x\geq\frac{4b-1}{4b}$} \\ [2mm]
\frac{l}{2b} & \mbox{if $\frac{2l-1}{4b}\leq x< \frac{2l+1}{4b}$, \quad $l=1,\dots,b-1$} \\ [2mm]
0 & \mbox{if $x<\frac{1}{4b}$}
\end{array} \right.
\]

\begin{example}
For $b=2$, a linear combination  
\[
    \overline{B}[(B_1,B_2),(1/2,1/2)]=\frac{1}{2}(B_1+B_2).
\]
of two bucket matrices $B_1,B_2$ with weights $w^1=w^2=1/2$, has its entries in the set $\{0,1/4,2/4,3/4,1\}$. Then, the second utopian matrix $U_C^2$ associated with the pair order matrix $C=\overline{B}[(B_1,B_2),(1/2,1/2)]$ becomes
\[
U_C^2(u,v)=\Upsilon^2(C(u,v))
\]
where
\[
\Upsilon^2(x)=\left\{
\begin{array}{rl}
1 \qquad &  \mbox{if $x \geq \frac{7}{8}$} \\ [2mm]
\frac{3}{4} \qquad &  \mbox{if $\frac{5}{8}\leq x < \frac{7}{8}$} \\ [2mm]
\frac{2}{4} \qquad &  \mbox{if $\frac{3}{8}\leq x < \frac{5}{8}$} \\ [2mm]
\frac{1}{4} \qquad &  \mbox{if $\frac{1}{8}\leq x < \frac{3}{8}$} \\ [2mm]
0 \qquad &  \mbox{if $x < \frac{1}{8}$}
\end{array}
\right.
\]
\end{example}

Analogously, we define 
the \emph{$b$-th utopia value} $u_C^2$ associated with $C$ as $u_C^2=D(U_C^2,C)$. 

If the utopian matrix $U_C^b$ can be expressed as a linear combination of $b$ bucket order matrices $B_1,\dots,B_b$ with weights $w^k=1/b$, then $\{B_i\}^b_1$ is an optimal solution to the OSBOP$_e^b$, i.e. 
$D(B,C)\geq u_C^b$ for whichever matrix $B$ which is a linear combination, with equal weights $1/b$, of $b$ bucket matrices.

\begin{remark}
It is worth noting that the $b$-th utopian matrix $U_C^b$ has been defined in the framework of the {\rm OSBOP}$^b_e$, where all the weights $w^k$ are equal to $1/b$. If this condition is not set, the definition of $b$-th utopian matrix does not apply.
\end{remark}

\begin{example} The dataset 4-2 taken from \cite{Mattei2013} with $n=3$ has the following associated pair order matrix $C$ 
\[
C=\left(
  \begin{array}{ccc}
0.5000     & 0.7046  &  0.4934 \\
0.2954     & 0.5000  &  0.3790  \\
0.5066     & 0.6210  &  0.5000     \\
  \end{array}
\right).
\]
The utopian matrices of orders $b=1,2,3,4$ of $C$ are, respectively\footnote{We have not simplify the quotients to make it easier to identify the values defined for the matrices $U_C^b$},

\[
U_C = \left(
  \begin{array}{ccc}
1/2 & 1/2   &  1/2 \\
1/2 & 1/2   &  1/2  \\
1/2 & 1/2   &  1/2   \\
  \end{array}
\right),
\quad 
U_C^2 = \left(
  \begin{array}{ccc}
2/4     & 3/4  &  2/4 \\
1/4     & 2/4  &  2/4 \\
2/4     & 2/4  &  2/4 \\
  \end{array}
\right),
\]

\[
U_C^3 = \left(
  \begin{array}{ccc}
3/6  & 4/6    &  3/6 \\
2/6  & 3/6    &  2/6  \\
3/6  & 4/6    &  3/6   \\
  \end{array}
\right), 
\quad
U_C^4 = \left(
  \begin{array}{cccc}
4/8     & 6/8   &  4/8 \\
2/8     & 4/8   &  3/8 \\
4/8     & 5/8  &  4/8 \\
  \end{array}
\right).
\quad 
\]

The associated utopia values are  $u_C=0.6644$,  $u_C^2=0.3460$, $u_C^3=0.1804$ and $u_C^4=0.112$. 

Observe that the utopian matrix $U_C$ for the {\rm OBOP} is feasible, i.e. is a bucket matrix associated with the bucket order with all the items $1,2,3$ tied. Consequently, $U_C$ is an (optimal) solution to the {\rm OBOP}, being 
\[
D(C,U_C)=u_C=0.6644.
\]

Similarly, the second utopian matrix $U_C^2$ for the {\rm OSBOP}$^2_e$ can be expressed as the linear combination of the two bucket orders $1|2|3$ and $3|1,2$, i.e.
\[
\left(
  \begin{array}{ccc}
2/4     & 3/4  &  2/4 \\
1/4     & 2/4  &  2/4 \\
2/4     & 2/4  &  2/4 \\
  \end{array}
\right)=\frac{1}{2}\left(
  \begin{array}{ccc}
1/2     & 1  &  1 \\
0     & 1/2  &  1 \\
0     & 0  &  1/2 \\
  \end{array}
\right)
+ \frac{1}{2}\left(
  \begin{array}{ccc}
1/2     & 1/2  &  0 \\
1/2     & 1/2  &  0 \\
1     & 1  &  1/2 \\
  \end{array}
\right).
\]
Namely, $U_C^2$ is an (optimal) solution to the {\rm OBOP}$_e^2$, being 
\[
D(C,U^2_C)=u_C^2=0.3460.
\]

Also the third utopian matrix $U_C^3$ for the {\rm OSBOP}$^3_e$ can be expressed as the linear combination of the three bucket orders  $1,2,3$, $1|2,3$ and $3|1,2$, i.e.
\begin{scriptsize}
\[
\left(
  \begin{array}{ccc}
3/6  & 4/6    &  3/6 \\
2/6  & 3/6    &  2/6  \\
3/6  & 4/6    &  3/6 \\
  \end{array}
\right)=\frac{1}{3}\left(
  \begin{array}{ccc}
1/2 & 1/2   &  1/2 \\
1/2 & 1/2   &  1/2  \\
1/2 & 1/2   &  1/2   \\
  \end{array}
\right)
+ \frac{1}{3}\left(
  \begin{array}{ccc}
1/2     & 1  &  1 \\
0     & 1/2  &  1/2 \\
0     & 1/2  &  1/2 \\
  \end{array}
\right)
+ \frac{1}{3}\left(
  \begin{array}{ccc}
1/2     & 1/2  &  0 \\
1/2     & 1/2  &  0 \\
1     & 1  &  1/2 \\
  \end{array}
\right),
\]
\end{scriptsize}
and so $U_C^3$ is an (optimal) solution to the {\rm OBOP}$_e^3$, being 
\[
D(C,U^3_C)=u_C^3=0.1804.
\]

Finally, the 4-th utopian matrix $U_C^4$ is also feasible.
In Table \ref{tab:BestSolutionsSmall} of Section \ref{ILL} this example is analyzed in more detail. In particular, there are three (optimal) solutions to the {\rm OSBOP}$_e^4$. As can be easily checked, for the three solutions the fourth utopian matrix $U_C^4$ is obtained as the linear combination (with weights equal to 1/4) of the respective bucket orders, being $D(C, U_C^4)=u_C^4=0.1120$.

This example illustrates the improvement (in terms of distance) that can be obtained by increasingly including more bucket orders in the solution.

To finish, it is important to note that, in some cases, the utopian matrix cannot be expressed as a linear combination of $b$ bucket matrices with equal weights  $1/b$. For example, in \cite{Aledo2017Utopia}, the precedence matrix associated with the set of rankings ${1|2|3, 1|3|2, 2|1|3}$ is presented as an instance that cannot be expressed using a single bucket order (i.e., {\rm OSBOP}$^1_e$). However, $U_C^3$ becomes feasible for the {\rm OSBOP}$_e^3$ by employing these three bucket orders with equal weights $1/3$.
\end{example}

\subsection{Illustrative examples of {\rm OSBOP} vs {\rm OBOP}}\label{ILL}

In  Table \ref{tab:BestSolutionsSmall} we display the solutions of several versions of the OSBOP for some small instances\footnote{Datasets taken from PrefLib \cite{Mattei2013}}. For the cases  of equal weights, the solutions were obtained by exhaustive search. 

Specifically, column Id is the label of the dataset, column Problem indicates the OSBOP variant, 
column $f_C(\overline{B})$ shows the distance from the best solution found to the input pair order matrix, and column Solution shows such solution(s) detailing both the bucket orders and the weights. When the rankings in the datasets have three elements (Id 4-1 and Id 4-2), we solve the OBOP, the OSBOP$_e^2$, the OSBOP$^2$, the OSBOP$_e^3$ and the OSBOP$_e^4$. 
When the rankings in the datasets have more than three elements (Id 2-1 and Id 2-2), we only solve the  OBOP, the OSBOP$_e^2$ and the OSBOP$^2$. For all the datasets, the OSBOP$_e$ with all the weights equal are solved by exhaustive exploration, so obtaining the (optimal) solution(s). Regarding the OSBOP$^2$, it is solved by using the metaheuristic presented in Section \ref{sec:method}, so showing  the best solution found (not guaranteed to be the optimal one).

\begin{table}
\begin{footnotesize}
\centering
\caption{\label{tab:BestSolutionsSmall}Best solutions obtained in small datasets}
\renewcommand{\arraystretch}{1.2} 
\smallskip
\begin{tabular}{llcl}\hline
Id	&	Problem &   $f_C(\overline{B})$    & Solution	\\  \hline
2-1	&	OBOP        &   1.4636  & $1,2,3|4$	\\ \cline{2-4}  
    &   OSBOP$^2_e$ &   0.9816  & (1/2) $\cdot$ $1|3|2,4$ + (1/2) $\cdot$ $2,3|1,4$ \\
     \cline{2-4}
    &   OSBOP$^2$   &   0.4216  & 0.8328 $\cdot$ $1,2,3|4$ + 0.1672 $\cdot$ $3,4|1|2$  \\
     \hline
2-2	&	OBOP        &   1.4303  & $2,3,4|1,5$	\\  \cline{2-4}
    &   OSBOP$^2_e$ &   1.1754  & (1/2) $\cdot$ $3,4|2|5|1$ + (1/2) $\cdot$ $2|1,3,4|5$ \\
    \cline{2-4}
    &   OSBOP$^2$   &   0.4586  &  0.8611 $\cdot$ $2,3,4|1,5$ + 0.1389 $\cdot$ $1|2,3,4,5$  \\
    \hline
4-1	&	OBOP        &   0.5783  & $1,2|3$	\\  \cline{2-4}
    &   OSBOP$^2_e$ &   0.4398  & (1/2) $\cdot$ $1|2,3$ + (1/2) $\cdot$ $2|1|3$ \\
    \cline{2-4}
    &   OSBOP$^2$   &   0.1325  & 0.7188 $\cdot$ $1,2|3$ + 0.2812 $\cdot$ $1,2,3$   \\
    \cline{2-4}
    &   OSBOP$^3_e$   &    0.1606       & (1/3) $\cdot$ $1|2,3$ + (1/3) $\cdot$ $1,2|3$ +(1/3)$\cdot$ $2|1,3$ 
    \\ \cline{4-4}    
    &  &     & (1/3) $\cdot$ $1|2|3$ + (1/3) $\cdot$ $1,2,3$ + (1/3) $\cdot$ $2|1|3$ \\ \cline{2-4}
    &   OSBOP$^4_e$ &   0.1325  & 
   
    (1/4) $\cdot$ $1|2|3$ + 
    (1/4) $\cdot$ $2|1|3$ + 
    (1/4) $\cdot$ $1,2|3$ + 
    (1/4) $\cdot$ $1,2,3$    \\ \cline{4-4}
    &   &     & 
   
    (1/4) $\cdot$ $1|2|3$ + 
    (1/4) $\cdot$ $2|1|3$ + 
    (1/4) $\cdot$ $2|1,3$ + 
    (1/4) $\cdot$ $1|2,3$      

    \\ \hline     
4-2	&	OBOP        &   0.6644  & $1,2,3$	\\  \cline{2-4}
    &   OSBOP$^2_e$ &   0.3460  & (1/2) $\cdot$ $3|1,2$ + (1/2) $\cdot$ $1|2|3$ \\ \cline{2-4}
    &   OSBOP$^2$   &   0.1804  & 0.7460 $\cdot$ $1,2,3$ + 0.2540 $\cdot$ $1,3|2$   \\ \cline{2-4}
    &   OSBOP$^3_e$ &   0.1804  & (1/3) $\cdot$ $1,2,3$ + (1/3) $\cdot$ $1|2,3$ + (1/3) $\cdot$ $3|1,2$ \\ \cline{4-4}
    &               &           & (1/3) $\cdot$ $1|2|3$ + (1/3) $\cdot$ $3|2|1$ + (1/3) $\cdot$ $1,3|2$ \\ \cline{4-4}
    &               &           & (1/3) $\cdot$ $1|3|2$ + (1/3) $\cdot$ $2|3|1$ + (1/3) $\cdot$ $1,3|2$ 
  \\ \cline{2-4}
    &   OSBOP$^4_e$ &    0.1120  & 
    (1/4) $\cdot$ $1|2|3$ + 
    (1/4) $\cdot$ $1|3|2$ + 
    (1/4) $\cdot$ $3|1|2$ + 
    (1/4) $\cdot$ $2,3|1$  \\ 
    \cline{4-4}
    &               &           & 
    (1/4) $\cdot$ $1|2|3$ + 
    (1/4) $\cdot$ $1|2,3$ + 
    (1/4) $\cdot$ $3|1|2$ + 
    (1/4) $\cdot$ $3|2|1$ 
     \\  \cline{4-4}
    &               &           & 
    (1/4) $\cdot$ $1|2|3$ + 
    (1/4) $\cdot$ $1,2,3$ + 
    (1/4) $\cdot$ $1,3|2$ + 
    (1/4) $\cdot$ $3|1,2$     
    \\ \hline

\end{tabular}
\end{footnotesize}
\end{table}

It is worth noting some interesting aspects:

\begin{itemize}   
    \item Although our experiments show that larger values of
$b$ tend to yield better solutions for OSBOP$^b_e$, this cannot be considered as a general fact. Nevertheless, they can be regarded as alternative approaches to interpreting the input matrix through a greater number of rankings.

    \item In the four datasets considered, the solution for the OSBOP$^2$ includes as the most important bucket order (i.e. greater weight) the bucket order  solution for the OBOP. In other words, the solutions for the OSBOP$^2$ are refinements of the unitary solutions of the OBOP which include precedence relations not considered in the latter.
   Actually, in some cases (Id 2-1 and Id 2-2) the second bucket order included in the solution is very different to the one with the greater weight. 
   It can be understood as a multimodal landscape of the input preferences, which may be viewed as the existence of different (minority) opinions with respect to the majority group.  
   
   For example, the second ranking in the solution for the dataset Id 2-1 is $3,4|1|2$, which differs in all the precedence relations with respect to  $1,2,3|4$. Similarly, the second ranking in  the dataset Id 2-2 is $1|2,3,4,5$, which differs with respect to the first ranking $2,3,4|1,5$ in most precedence relations, except for those among $2$, $3$ and $4$, which are considered tied in both rankings. In both cases, although the weight of the principal (first) ranking is remarkably high  (0.83 and 0.86, respectively), the improvement of the solution to the OSBOP$^2$ with respect to the one of the OBOP is outstanding (71\% and 68\%, respectively). 
   On the other hand, in the datasets Id 4-1 and Id 4-2 the second bucket order is quite similar to the first one: $1,2,3$ and $1,2|3$ in dataset Id 4-1, and $1,3|2$ and $1,2,3$ in dataset Id 4-2. However, the improvement with respect to the solution to the OBOP is even more noteworthy (77\% and 73\%, respectively).
   
   \item Contrary to what was described in the previous paragraph, in the four datasets considered the solution to the OSBOP$^2_e$  does not contain any bucket order present in either the OBOP solution or the OSBOP$^2$ solution.
    This shows that they are largely distinct consensus approachings. For example, in dataset Id 2-1 the solution to the OSBOP$^2_e$ (forced to consist of two equally weighted rankings) includes the bucket orders $1|3|2,4$ and $2,3|1,4$, which significantly differs from the solution to the OSBOP$^2$ consisting of $1,2,3|4$ (with weight 0.8328) and $3,4|1|2$ (with weight 0.1632). 
   
    \item As suggested by the instances analyzed, the flexibility of the weights (i.e. when they are not forced to be equal)  appears to have a greater impact on the solution than the number of bucket order matrices considered ($b$). In fact, in the datasets Id 4-1 and Id 4-2 the solution found to the OSBOP$^2$ is better than the (optimal) one to the OSBOP$^3_e$.

\end{itemize}

All in all, from a decision maker point of view and depending on the problem, it may be assessed whether the improvement in the distance justifies the inclusion of additional rankings in the solution. In other words, the principle of Occam's razor may be considered.

\section{A simple metaheuristic approach for OSBOP} \label{sec:method}

In this section we present SLS-OSBOP, a stochastic local search procedure  to solve  the OSBOP in the case where the exhaustive exploration is computationally intractable.
The search is structured in two levels. In an outer level, the method searches for a vector of $b$ bucket orders ($s_c$), while in an inner level a numerical optimization procedure is carried out to tune the weights ($w_c$) for the bucket orders in $s_c$.

The method, outlined in Figure \ref{fig:lsobopf},  takes as input the precedence matrix $C$, the number $b$ of desired rankings (bucket orders) in the output, a maximum number of iterations $t_1$ for the bucket orders search, a maximum number of iterations $t_2$ for the weights tuning procedure, and the binary variable $Eq$ to set whether all the rankings in the output have equal weight ($Eq$=yes) or if different weights are allowed ($Eq$=no).

SLS-OSBOP firstly generates an initial (random) candidate solution $s_c$  and takes a vector of weights $w_c$ with equal value $1/b$ for all the rankings in the solution (Steps 1-2). Next, the candidate solution $s_c$ is evaluated (Step 3). 
Then, if $Eq$=no,  an inner local search procedure {\sc TuneWeights}($C$,$s_c$,$w_c$,$f_c$,$t_2$) is executed to carry out random changes in the weights, returning those that produce the best (minimum) evaluation value. The best vector of weights obtained is taken as the weight $w_c$ of the current solution $s_c$ and the evaluation $f_c$ is updated (Step 4).

Subsequently, the main (outer) search iterates $t_1$ times  to improve the current solution $s_c$ (Steps 5-15). In order to do so, in each iteration a new solution $s_c'$ is generated by mutating $s_c$ (Step 6). This solution $s_c'$ undergoes in Steps 6-8 the same procedure carried out by  $s_c$ in Steps 1-3 in order to assign the corresponding weights and to evaluate its quality. 
If the new solution $s_c'$ is better or equal than the current solution $s_c$, then $s_c$ is updated to $s_c'$ (Steps 11-13). We accept solutions with equal $f_c$ to allow to visit more regions of the search space. Finally, the best solution obtained $(s_c,w_c)$ is returned jointly with its evaluation $f_c$.

\begin{figure}[htbp]\label{tab:lsosbop}
\centering
\begin{tabular}{ll}\hline
\multicolumn{2}{c}{$\mathbf{SLS-OSBOP}(C, b, Eq, t)$} \\ \hline
\multicolumn{2}{l}{{\bf Input}: $C$, precedence matrix; 
 $b$, number of buckets in the output; 
 $Eq\in \{ \text{yes},\text{no}\}$;}\\
\multicolumn{2}{l}{\hspace*{1.1cm} $t_1$, iterations for the outer loop; $t_2$, iterations for the numerical optimization}\\

\multicolumn{2}{l}{{\bf Output}: Set of bucket orders}\\
1&  $s_c \leftarrow$ {\sc InitialSolution}$(C,b)$ \\
2&  $w_c \leftarrow (1/b,\dots,1/b)$ \\
3&  $f_c \leftarrow$ {\sc EvaluateSolution}$(C,s_c,w_c)$ \\
4 & \textbf{if} $Eq$=no  \textbf{then} $(f_c,w_c) \leftarrow$ {\sc TuneWeights}$(C,s_c,w_c,f_c,t_2)$   \\
5 & \textbf{for} $i=1$ to $t_1$ \textbf{do} \\
6 & \hspace{0.5cm} $s_c' \leftarrow$ {\sc MutateSolution}$(s_c)$ \\
7&  \hspace{0.5cm} $w_c' \leftarrow (1/b,\dots,1/b)$ \\
8&  \hspace{0.5cm} $f_c' \leftarrow$ {\sc EvaluateSolution}$(C,s_c',w_c')$ \\
9 & \hspace{0.5cm} \textbf{if} $Eq$=no  \textbf{then} $(f_c',w_c') \leftarrow$    {\sc TuneWeights}$(C,s_c',w_c',f_c',t_2)$ \\
10 & \hspace{0.5cm} \textbf{if} $f_c' \leq f_c$ \textbf{then} \\
11 & \hspace{1.0cm} $s_c \leftarrow s_c'$  \\
12 & \hspace{1.0cm} $w_c \leftarrow w_c'$  \\
13 & \hspace{1.0cm} $f_c \leftarrow f_c'$  \\
14 & \textbf{end for} \\
15 & \textbf{return} $s_c$, $w_c$, $f_c$ \\ \hline
\end{tabular}
\caption{A local search approach for OSBOP}
\label{fig:lsobopf}
\end{figure}

Let us briefly describe the functions involved in SLS-OSBOP.

The method {\sc InitialSolution}$(C,b)$ randomly generates a vector $s_c$ of $b$ bucket orders. 
Note that other heuristic procedures may be used to generate $s_c$, e.g. using as one of the bucket orders the solution to the OBOP.

For each solution $s_c$ with $b$ bucket orders $\mathcal{B}_1,\dots \mathcal{B}_b$ (i.e. $b$ bucket order matrices $B_1,\dots,B_b$)  and each vector of weights $w_c=(w^1,\dots,w^b)$, we construct the  matrix $ \overline{B}[(B_1,\dots,B_b),(w^1,\dots,w^b)]$ as in (\ref{Bbar}), and use the function $f_C$ (\ref{fC}) to evaluate the solution
\begin{equation}\label{ES}
\text{{\sc EvaluateSolution}}(C,s_c,w_c)=f_C( \overline{B}[(B_1,\dots,B_b),(w^1,\dots,w^b)],C)
\end{equation}

The local search procedure {\sc TuneWeights}$(C,s_c,w_c,f_c,t_2)$ is executed to find the weights  that minimize (\ref{ES}).  Specifically, a local search procedure is applied to improve the objective function. Starting from equal weights $1/b$ for  the $b$ bucket orders, a sequence of random changes\footnote{We have set the number of changes to 100 in our experimentation} is applied. In particular, a random value $r \in [-0.5,0.5]$ is generated for a random weight $w$ so the new value becomes  
$max(min(w+r,1),0)\in [0,1]$. Then, the values of the remaining weights are proportionally adjusted to ensure that their total equals 1. If the value of the objective function is improved with  the change, then it is accepted; otherwise the change is discarded and the previous weights are kept.

The method {\sc MutateSolution}($s_c$) takes two decisions at random: the bucket order to be mutated and the mutation to be applied. Also the number of bucket orders  to be mutated is chosen at random. For  each bucket order to be mutated, any of the mutations introduced in \cite{Aledo2018-ES-EJOR}  may be applied: 
\begin{itemize}
\item Bucket Insertion: Randomly pick a bucket and place it in another random position.
Example: $8,7|6|5,4,3|\mathbf{2,1}\longrightarrow\mathbf{2,1}|8,7|6|5,4,3$
\item Buckets Interchange: Randomly pick two buckets and interchange their positions.
Example: $\mathbf{8,7}|6|5,4,3|\mathbf{2,1}\longrightarrow\mathbf{2,1}|6|5,4,3|\mathbf{8,7}$
\item Bucket Inversion: Randomly pick a chain of consecutive buckets and reverse the order of the buckets.
Example: $\mathbf{8,7|6|5,4,3|}2,1\longrightarrow\mathbf{5,4,3|6|8,7|}2,1$
\item Bucket Union: Randomly pick two consecutive buckets and join them.
Example: $\mathbf{8,7|6}|5,4,3|2,1$ $\longrightarrow \mathbf{8,7,6}|5,4,3|2,1$
\item Bucket division: Randomly pick a bucket with two or more items and split it into two consecutive buckets. Each item is randomly placed in one of the resulting buckets.
Example: $8,7|6|\mathbf{5,4,3}|2,1\longrightarrow
8,7|6|\mathbf{5,3|4}|2,1$
\item Item Insertion: Randomly pick a bucket and one of its items. Then either the item is inserted in an existing bucket or generates a new (singleton) bucket. Both decisions are randomly taken.
Example: $8,7|6|\mathbf{5},4,3|2,1\longrightarrow
8,7,\mathbf{5}|6|4,3|2,1$
\item Item Interchange: Randomly pick two buckets and one item of each bucket, and interchange them. Example:
$8,\mathbf{7}|6|\mathbf{5},4,3|2,1\longrightarrow
8,\mathbf{5}|6|\mathbf{7},4,3|2,1$
\end{itemize}

The first three mutations are frequently used in permutation problems \cite{Ceberio:TEC:2014,Talbi2009} and they do not produce any change in the set of buckets, just in their order. 
The other mutations modify the set of buckets, either their number or/and their internal composition.

\section{Experimental results}\label{sec:experiments}

This section presents an experimental study using 14 datasets from PrefLib \cite{Mattei2013}, each with fewer than 25 items to be ranked ($n<25$), since in this seminal work we are interested in interpreting the results and comparing alternative solutions to different variants of the OSBOP. 
Specifically, we considered three  OSBOP variants: OBOP ($b=1$), OSBOP$^2_e$ ($b=2$ and $Eq=$yes), and OSBOP$^2$ ($b=2$ and $Eq=$no).
Our aim is to analyze how increasing the model's flexibility (i.e., allowing more bucket orders in the solution and non-equal weights) enhances the accuracy of the response in describing the input matrix (i.e., reducing the distance between the solution and the input matrix). For the experiments, we set $t_1=10000$ and $t_2=100$.

\subsection{Results}

 The implementation of the algorithms, provided to ensure the reproducibility of the results, is available at \url{https://swish.swi-prolog.org/p/osbop.pl}. Table \ref{tab:ResultsOSBOP} presents the best results achieved for the OBOP, the  OSBOP$^2_e$  and the OSBOP$^2$ over the 14 datasets, whose identifiers are specified in the column Dataset-Id. For each dataset, we show  the number of items ($n$), and the utopia values $u_C$ and $u_C^2$. The values in columns OBOP, OSBOP$^2_e$ and OSBOP$^2$ correspond to the distance of the best solution to the input matrix obtained after $t_1=10000$ iterations. The column $w_1$ shows the weight of the ranking with the highest weight in the best solution found for OSBOP$^2$.

\begin{table}
\centering
\caption{\label{tab:ResultsOSBOP}Results of OBOP vs. OSBOP in the selected benchmark of datasets.}
\renewcommand{\arraystretch}{1.2} 
\smallskip
\begin{tabular}{lrrrrrrr}\hline
Dataset-Id	&	$n$ &	    $u_C$&$u^2_C$      &	OBOP    &	OSBOP$^2_e$ & OSBOP$^2$	&	$w_1$ \\\hline
2-1	&	4 &		 1.46&0.98  &	 1.46	&   0.98    &   0.42	&   0.83\\   
2-2	&	5 &	     1.43&1.18  &    1.43   &   1.18    &   0.46    &   0.86\\
4-1	&	3 &		 0.58&0.44  &	 0.58	&   0.44    &   0.13	&   0.72\\   
4-2	&	3 &	     0.66&0.35  &    0.66   &   0.35    &   0.18    &   0.75\\
6-3	&	14 &	 5.00&2.39	&    5.67	&   2.89	&   2.89	&   0.50\\ 
6-4	&	14 &	 2.33&1.39	&    2.67	&   1.44    &   1.44 	&   0.50\\ 
6-11 &	20 &	12.67&7.11	&   14.22	&   8.11    &   6.67 	&    0.67\\
6-12 &	20 &	 5.67&4.39	&    5.67	&   4.61    &   4.19 	&   0.57\\ 
6-18 &	24 &	 7.33&4.00	    &    7.67	&   4.00    &   3.28	&   0.55\\ 
6-28 &	24 &	24.22&12.89	&   30.33	&  16.94    &  15.44 	&   0.67\\
6-48 &	24 &	10.67&4.33	&   12.11   &   6.17    &   5.56    &  0.56 \\ 
14-1 &	10 &	11.69&4.99 	&   13.09 	&   6.73    &   5.25 	&   0.59\\
15-48 &	10 &	 9.33&4.67	&   13.00   &   7.83    &   4.67 	&   0.67\\ 
15-74 &	20 &	26.33&13.17	&   40.00	&  26.00    &  14.67 	&   0.67\\ \hline
\end{tabular}
\end{table}

Taking into account the solutions obtained for the 14 datasets, the average ratio OSBOP$^2_e$/OBOP is 0.61, the average ratio OSBOP$^2$/OBOP is 0.42 and the average ratio OSBOP$^2$/ OSBOP$^2_e$ is 0.71. On the other hand, note that in the datasets Id 2-1, Id 2-2, Id 4-1, Id 4-2 and Id 6-12, the evaluation value ($f_c$) of the solution found for the OBOP coincides with the utopia value, that is, is, the solution found is optimal. Recall that the utopia value is a lower bound for the solution of OBOP that is not always feasible (see Section \ref{utopian matrix OSBOP}). In our experimentation, the average ratio OBOP/$u_C$ is 1.13. Although $u_C$ is neither a lower bound for OSBOP$^2_e$ nor for OSBOP$^2$, the average ratios  OSBOP$^2_e$/$u_C$ (0.69) and OSBOP$^2$/$u_C$ (0.48) also illustrate to what extent the fitness of the solution improves when a greater flexibility is allowed in the output. 
The average ratios OSBOP$^2_e$/$u^2_C$ (1.23) and OSBOP$^2$/$u^2_C$ (0.87) show the additional flexibility introduced by allowing different weights.
Recall that when the evaluation value ($f_c$) of the solution found for OSBOP$^2_e$ coincides with $u_C^2$, the solution is optimal (datasets Id 2-1, Id 2-2, Id 4-1, Id 4-2 and Id 6-18). 
Note also that the optimal utopian condition assumes equal weights, that is, $u^2_C$ is not a lower bound for OSBOP$^2$.

Although the OBOP is usually addressed by means of greedy algorithms \cite{Aledo2017Utopia,Ukkonen2009,Gionis2006}, in this study we are solving the OBOP with the heuristic SLS-OSBOP ($b=1$) for a fair comparison\footnote{Note that the results obtained using SLS-OSBOP are equal or better than those reported in \cite{Aledo2017Utopia} by means of greedy methods.} with the rest of OSBOP variants.

It is worth noting that, since the same number of iterations $t_1=10000$ was applied to all the OSBOP variants, the percentage of the search space explored when solving the OBOP is significantly larger than that explored when solving the OSBOP$^2_e$ and the OSBOP$^2$. Nonetheless, the results obtained for the OSBOP$^2_e$ and the OSBOP$^2$ provide a much closer approximation to the input preference matrix.

\subsection{Best solutions found for the OSBOP}

In Table \ref{tab:BestSolutions} we show the best solution found for the OBOP, the OSBOP$^2_e$ and the OSBOP$^2$ over the 10 largest datasets considered. Although already shown in Table \ref{tab:ResultsOSBOP}, for each dataset and OSBOP variant, we display the value $f_C(\overline{B})$, that is, the distance from the solution to the input matrix $C$ and, below each OSBOP$^2$ instance, the value $w^1$ corresponding to the bucket order with the highest weight ($w^2=1-w^1$, not shown). In particular, $w^1=w^2=0.5$ for the OSBOP$^2_e$.   
When the best solution obtained for the OSBOP$^2$ has $w^1=0.5$, it is also the solution to the OSBOP$^2_e$. In these cases (datasets with Id 6-3 and Id 6-4) this solution is only shown once. 
In some of these datasets there are different solutions with the same evaluation value ($f_C$). In these cases, Table \ref{tab:BestSolutions} shows only one of these solutions with equal score;
however, in practice, all of them should be presented to the practitioner for decision-making.

\begin{table}
\caption{\label{tab:BestSolutions}Best solutions found in the selected benchmark of datasets.}
\renewcommand{\arraystretch}{1.2} 
\smallskip
\begin{scriptsize}
\centering
\begin{tabular}{llcl}\hline
Id	&	Problem &   $f_C(\overline{B})$       & Solution 	\\  \hline
6-3	&	OBOP    &   5.6667  &       $10|7|5|8|2,13|1|4,11|14|6|9|3,12$	\\  \cline{2-4}
      &   OSBOP$^2$   &   2.8889 &    $10|7|5|8,13|2|1,4,11|6,14|3,9,12$  \\
    &   0.5      &          &       $10|7|5|8|2|13|1|11|4,14|6,9|12|3$  \\\hline
    
6-4	&	OBOP    &   2.6667  &       $11|14|12|13|9|10|7|8|5|6|4|3|2|1$	\\  \cline{2-4}
    &   OSBOP$^2$   &   1.4444 &    $11|14|12|9,13|10|7|8|5|6|4|3|2|1$  \\
    &   0.5      &          &       $11|14|12|13|9|10|5,7,8|6|4|3|2|1$  \\\hline
    
6-11	&	OBOP    & 14.22 &       $12|8|14|17|2,10,11,13|9,16|1|6,19|5,15,20|3|4,18|7$	\\  \cline{2-4}
    &   OSBOP$^2_e$   &   8.1111 &    $8,12,14,17|2,10,11,13|16|1,9|19|5,6,20|15|3,18|4|7$  \\
    &   0.5      &          &       $12|8|14|2,10,11,17|13|9,16|1,6|15,19,20|5|3|4|18|7$  \\ \cline{2-4}
    &   OSBOP$^2$   &   6.6692    &   $12|8|14, 17|11|13, 10, 2|16, 9|1]|19|5, 6, 20|15|3|4, 18|7$  \\
    &  0.6665       &          &      $12, 8, 14|10, 2|11|17, 13|16|6, 15, 1, 9|19, 20|5, 18|4, 3|7$  \\    
       
    \hline
    
6-12	&	OBOP  &  5.6667 &       $20|17,18|19|13,15,16|10|7,14|9,12|6,8|5|11|2|1|4|3$	\\  \cline{2-4}
    &   OSBOP$^2_e$  &   4.6111 &   $20|18|17|19|13|16|15|10|7,14|12|9|8|6|5|11|2|1|4|3$ \\
    &   0.5      &          &       $20|17|18|19|15,16|13|7,10,14|9|12|6,8|5|11|1,2|4|3$ \\ \cline{2-4}
    &   OSBOP$^2$   &   4.1877 &    $20|18|17|19|13|16|15|10|7,14|12|9|8|6|5|11|2|1|4|3$  \\
    &   0.5665   &          &       $20|17|18|19|15,16|13|7,10,14|9|12|6,8|5|11|1,2|4|3$  \\\hline
    
6-18	&	OBOP &   7.6667 &       $22|23|20,21|24|17,19|16|15,18|10,14|13|7,11,12|9|6,8|3,5|4|1|2$	\\  \cline{2-4}
    &   OSBOP$^2_e$  &  4       &   $22|23|21|20|24|17,19|16|15,18|14|10,11,13|7,12|9|8|6|3,5|4|1|2$  \\
    &   0.5        &        &       $22|23|20|21|24|17|19|16|15|18|10|14|13|12|7|11|9|3,6,8|4,5|1,2$ \\ \cline{2-4}
    &   OSBOP$^2$   &   3.279  &    $22|23|21|20|24|17,19|16|15,18|14|10,11,13|7,12|9|8|6|3,5|4|1|2$  \\
    &   0.5515   &          &       $22|23|20|21|24|17|19|16|15|18|10|14|13|12|7|11|9|3,6,8|4,5|1,2$  \\\hline
    
6-28 &	OBOP    &   30.3333   &     $24|20|19,23|14,16|13,15,21|4,9,12,18,22|8|2,5,6,11,17|1,3,7,10$	\\  \cline{2-4}
    &   OSBOP$^2_e$  &   16.9444  & $24|20|19,23|16|14,15|4,9,13,18,21,22|6,8,11,12,17|2,5|3,7,10|1$ \\
    &   0.5    &           &        $20,24|19|14,21,23|13,15,16|9,12,22|18|4|2,6,8|1,3,5,10,11,17|7$ \\ \cline{2-4}
    &   OSBOP$^2$   &   15.4444   & $24|20|19,23|13,14,15,16,21|4,9,12,18,22|6,8,17|2,5,10,11|1,3,7$  \\
    &   0.6667        &       &     $20,24|19|14,23|16|15|9,22|21|13,18|4,8|2,3,6,11,12|1|5,7|10,17$  \\\hline
    
6-48 &	OBOP    &   12.1111   &       $20|22,24|23|15,21|19|17,18|16|11,12,13,14|10|8,9|5,6|1,2,3,4|7$	\\  \cline{2-4}
    &   OSBOP$^2_e$  &   6.1667   & $20,24|22|23|21|15|17,18,19|16|14|12|10,11,13|9|8|6|5|3,4|1,2,7$ \\
    &   0.5    &           &        $20|22|24|15,23|21|19|18|16,17|11|13|10,12,14|8,9|5,6|2|4|1|3,7$ \\ \cline{2-4}
    &   OSBOP$^2$   &   5.5606   &     $20|22|24|15,23|21|19|18|16,17|11|13|10,12,14|8,9|5,6|2|4|1|3,7$
      \\
    &   0.5551    &         &    $20,24|22|23|21|15|17,18,19|16|14|12|10,11,13|9|8|6|5|3,4|1,2,7$     \\\hline
    
14-1 &	OBOP    &   13.0885   &     $1,2,3,4,5,6,7,8,10|9$	\\  \cline{2-4}
    &   OSBOP$^2_e$  &   6.7255   & $4,7|2,6,10|3,8|1,5,9$ \\
    &   0.5    &           &        $1,5|2,3,7,8,9,10|6|4$ \\ \cline{2-4}
    &   OSBOP$^2$   &   5.2501   &  $1,2,4,5,7,8,10|3,6,9$  \\
    &   0.5920        &       &     $3,6,7|1,2,4,5,8,9,10$  \\\hline
    
15-48 &	OBOP    &   13   &          $1|2|3,4,5,6,7,8,9|10$	\\  \cline{2-4}
    &   OSBOP$^2_e$  &   7.8333   & $1,4|2,6|3,5|7,8,9|10$ \\
    &   0.5    &           &        $7|1,8|2,9|3|4,5|6|10$ \\ \cline{2-4}
    &   OSBOP$^2$   &   4.6667   &  $1|2|3,4,5,7,8,9|6|10$  \\
    &   0.6667        &       &     $4|1|6|3|5|7|8|9|2|10$  \\\hline
    
15-74 &	OBOP    &   40   &          $2|1,3,4,5,6|7|8,9,10,11,12,13,14,16,17,20|15,18|19$	\\  \cline{2-4}
    &   OSBOP$^2_e$  &   26   &     $1,2|3|4,5,6|7,8,16|11,12,20|9,10,13,14,15,17|18|19$ \\
    &   0.5    &           &        $2,6|3,4,5|1,9|10|11|7,12,13|14|17|8,15,18,20|16,19$ \\ \cline{2-4}
    &   OSBOP$^2$   &   14.6667   & $2|3|1,4,5,6|7,9|8,10,11,12,17,20|13,14,15,18|16,19$  \\
    &   0.6667        &       &     $6|2|4|5|1|16|11|13|3|8|10|7|14|12|9|15|17|18|19|20$  \\\hline
\end{tabular}
\end{scriptsize}
\end{table}

\subsection{Discussion}

Next, we discuss some aspects of the solutions presented in Tables \ref{tab:ResultsOSBOP} and \ref{tab:BestSolutions}.
It is important to take into account that a stochastic greedy algorithm is used to solve the OSBOP instances, and therefore, the solutions obtained are unlikely to be optimal. However, they allow us to extract some general conclusions:

\begin{itemize}
    \item Only in two datasets (Id 6-3 and Id 6-4) the solutions  for OSBOP$^2$ and  OSBOP$^2_e$ coincide, i.e. only in these cases the flexibility of adjusting the weights did not produce a decreasing in the distance to the input precedence matrix.  
    \item There are three datasets (Id 6-12, Id 6-18 and Id 6-48) where the same pair of rankings are included in the best solution  to the OSBOP$^2_e$ and the OSBOP$^2$, but with different weights. These instances highlight how allowing flexibility in the weights leads to better solutions, so discovering communities with different importance (size). Specifically, there is an improvement of  12,4\% on average. In contrast, there are four datasets (Id 6-28, Id 14-1, Id 15-48 and Id 15-74) where the solutions for OSBOP$^2_e$ and OSBOP$^2$ have no ranking in common.
    \item It is  remarkable that the ranking found as a solution for the OBOP is never present in the pair of rankings 
 found as solution to the OSBOP$^2_e$ and the OSBOP$^2$. This supports the idea that considering two rankings in the response means a totally different approach to approximating the input matrix. 
    
\end{itemize}


\section{Conclusions}\label{sec:conclusions}

In this paper, we present a more general framework for addressing rank aggregation problems by considering a set of (weighted) rankings as the output, which more effectively captures the diversity present in the input data or within the search landscape. This approach allows the proposed ranking aggregation method to recognize the existence of distinct groups within the voter set or their expressed preferences, incorporating a fairness perspective that ensures diverse priorities and viewpoints are appropriately reflected in the aggregation process.

To illustrate our proposal, we define de Optimal Set of Bucket Orders Problem (OSBOP), a generalization of the Optimal Bucket Order Problem (OBOP), and provide insight through several illustrative examples. We also discuss the critical factors a decision maker must consider when choosing the variant of the OSBOP that best aligns with their specific needs and objectives. Finally, we introduce a simple metaheuristic approach for addressing the OSBOP and conduct an experimental study which shows that the proposed models yield solutions significantly closer to the input precedence matrix compared to those obtained when limiting the output to a single ranking (OBOP).

In future research, we plan to adapt specific heuristics from the OBOP for their application to OSBOP instances, as well as to design more complex metaheuristics to approach the OSBOP.



\section*{Acknowledgements}

This work has been partially funded by the Government of Castilla-La Mancha and ``ERDF A way of making Europe'' under project SBPLY/21/180225/000062; by MCIU/AEI/10.13039/501100011033 and ``ESF Investing your future'' through project PID2022-139293NB-C32; and by the Universidad de Castilla-La Mancha and ``ERDF A Way of Making Europe'' under project 2022-GRIN-34437.







\end{document}